\documentclass[letterpaper, 10 pt, conference]{ieeeconf}  % Comment this line out if you need a4paper

\IEEEoverridecommandlockouts                              % This command is only needed if 
\usepackage{graphicx}
\usepackage{amsmath} 
\usepackage{mathtools}
\usepackage{booktabs}
\usepackage{multirow}
\usepackage{dblfloatfix}
\title{\LARGE \bf
VecTile: A Low-Cost Flexible Tactile Array for 3-DoF Force Sensing
}

\author{Jian Hou,~\IEEEmembership{Student Member,~IEEE}, Xin Zhou,~\IEEEmembership{Member,~IEEE}, and Adam J. Spiers,~\IEEEmembership{Member,~IEEE}
\thanks{All authors are with the Manipulation and Touch Lab, Dept. Electrical and Electronic Engineering, Imperial College London. (Corresponding Author: jian.hou22@imperial.ac.uk)}
}

\begin{document}

\maketitle
\thispagestyle{empty}
\pagestyle{empty}

%%%%%%%%%%%%%%%%%%%%%%%%%%%%%%%%%%%%%%%%%%%%%%%%%%%%%%%%%%%%%%%%%%%%%%%%%%%%%%%%

\begin{abstract}
Tactile sensing provides robots with contact information difficult to obtain from vision alone.
Shear force is particularly important for understanding tangential contact, guiding corrective motions, and maintaining grasp stability, yet flexible and low-cost tactile arrays often measure only normal force.
We introduce VecTile, a low-cost, 6 mm thin, flexible tactile array capable of three-axis force sensing.
VecTile combines a piezoresistive layer with passive silicone nubs that convert normal and shear forces into distinguishable pressure patterns.
The sensor can be fabricated in-house without specialized equipment. The complete system costs less than 35 USD for a 4$\times$4 array.
We validate the sensing mechanism through finite element analysis and hardware experiments.
In controlled single-taxel tests over 1--5 N normal force and up to 1.5 N shear force, VecTile achieves root-mean-square errors of 0.39 N and 0.16 N for normal and shear forces. 
% Furthermore, we show that a VecTile calibrated in a flat configuration can be used in a curved configuration with minimal accuracy loss, and that a manual calibration sequence only lasting around 5 minutes is sufficient for practical use.
A VecTile calibrated while flat retains accuracy when curved, while a five-minute manual calibration is sufficient for practical use. 
Two applications demonstrate its utility. First, shear readings are sensitive enough to stop a sharp scalpel from puncturing a balloon after contact. Second, the sensor substantially improves Diffusion Policy-guided power-plug insertion. All designs and code will be released upon paper acceptance. Project website: https://jiannnh.github.io/VecTile/. 

\end{abstract}

%%%%%%%%%%%%%%%%%%%%%%%%%%%%%%%%%%%%%%%%%%%%%%%%%%%%%%%%%%%%%%%%%%%%%%%%%%%%%%%%

\section{Introduction}

Tactile sensing gives robots a sense of touch during manipulation. It reveals force, friction, stiffness, and other properties that are difficult to infer from vision alone \cite{meribout2024tactile, luo2025tactile}. This information is particularly important for grasp-stability estimation \cite{grover2022learning,saloutos2023towards}, object recognition \cite{spiers2016single,zhou2023tactile}, and in-hand manipulation \cite{yin2025learning,yin2023rotating,qi2023general}.

During manipulation, the normal-force component captures only part of the contact state. While the shear components reveal the direction and magnitude of tangential loading, supporting grasp-stability assessment, friction and slip reasoning, and contact regulation \cite{pattabiraman2025eflesh,ford2025shear}. Flexibility is also important because conformable tactile skins can cover non-planar and adaptive surfaces, including rounded fingers and soft grippers \cite{wistreich2025dexskin,huang2026flexitac}. Because direct contact exposes tactile sensors to wear and damage, low cost and repairability are important for practical deployment~\cite{hou2024location,bhirangi2021reskin}.

These considerations motivate tactile skins that combine 3-degree-of-freedom (DoF) force sensing, flexibility, and low cost. Existing tactile architectures, however, trade among these properties due to fabrication and integration complexities \cite{luo2025tactile}: Vision-based \cite{ford2025shear, yuan2017gelsight,lambeta2020digit} and magnetic-based \cite{pattabiraman2025eflesh, bhirangi2021reskin} sensors can recover rich multi-axis contact information but are rigid, while thin capacitive \cite{wistreich2025dexskin} and piezoresistive \cite{huang2026flexitac} arrays more readily conform to curved surfaces yet do not provide shear force information. An easily accessible flexible tactile array providing local 3-DoF force measurements therefore remains underexplored.

\begin{figure}[t]
    \centering
    \includegraphics[width=0.95\linewidth]{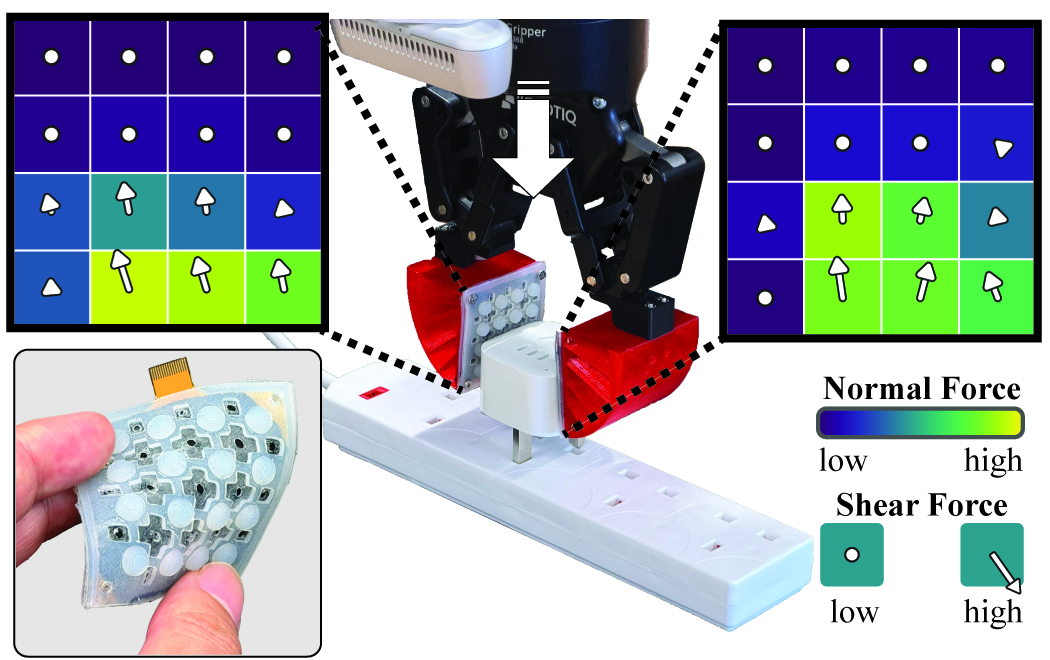}
    \caption{VecTile on Fin-Ray type fingers during plug insertion. Color denotes normal response and arrows indicate shear magnitude and direction.}
    \label{fig:teaser}
\end{figure}

We introduce VecTile (Fig.~\ref{fig:teaser}), named as such due to its capability of reading a 3-DoF force vector at each tactile pixel (taxel). VecTile is a low-cost, flexible, $6~\mathrm{mm}$-thin, $4\times4$ tactile array (16 local 3-DoF taxels at a $10~\mathrm{mm}$ pitch) that combines a piezoresistive array with passive silicone nubs. 

The complete sensor and readout board cost less than 35 USD (of which 25 USD is spent on an Arduino Nano that can be replaced with a lower-cost microcontroller and ADC circuit if needed). Raw frames stream at 343 Hz, and the measured end-to-end latency relative to a commercial force/torque (F/T) reference is approximately $25~\mathrm{ms}$. Most conveniently, this sensor can be easily produced in-house without the need for expensive equipment or tools. At its core is a combination of a standard flexible PCB (FPC), whose fabrication can be out-sourced at low cost (we used JLCPCB's services) and silicone parts that can be easily produced via 3D printed molds (we used PLA molds from a Bambu Lab P1S) and hobby-grade vacuum chamber (less than 150 USD). 
% All design files and code will be open-sourced upon publication.

\begin{figure*}[b!]
    \centering
    \includegraphics[width=0.9\linewidth]{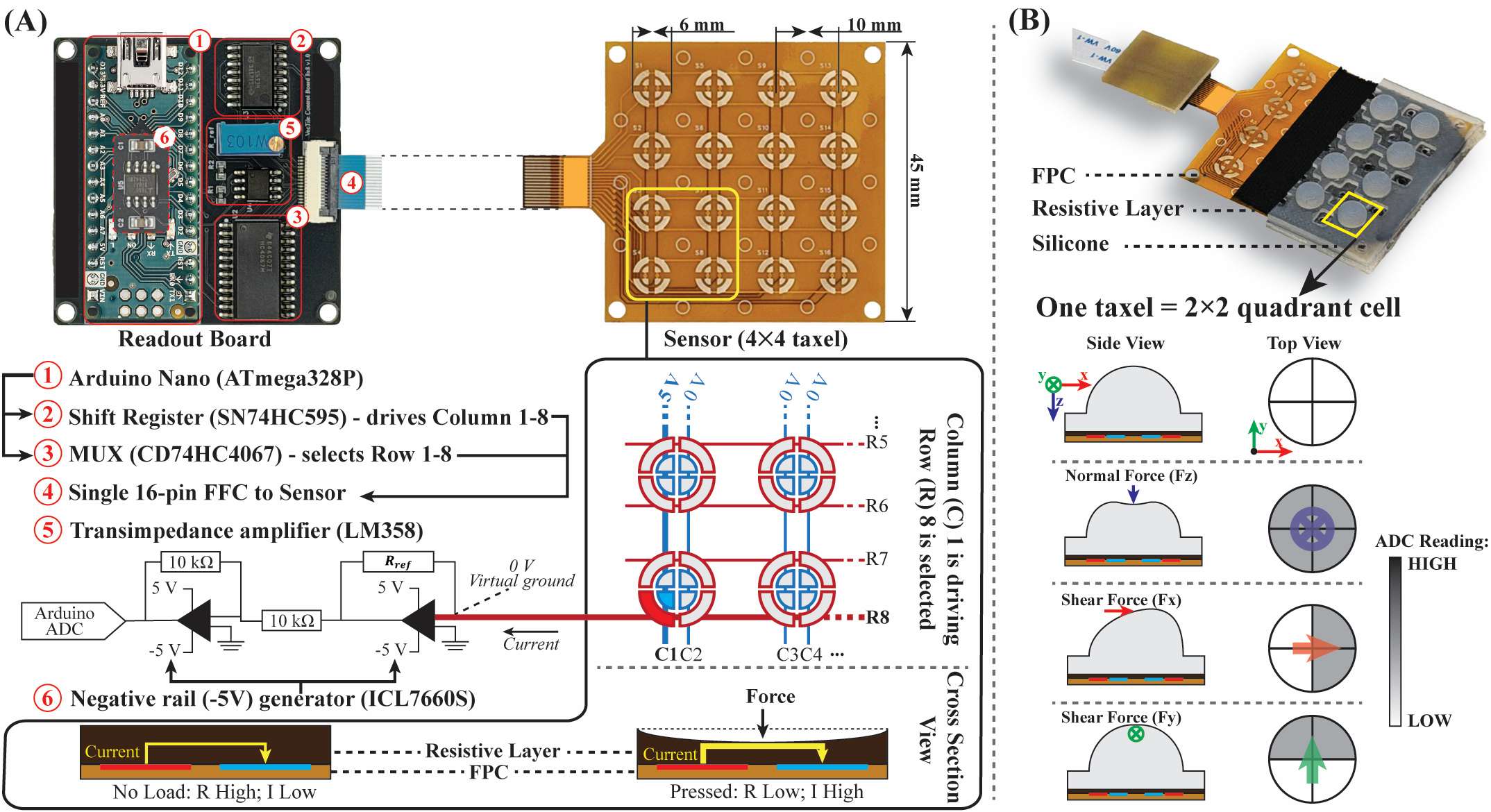}
    \caption{VecTile design.
    (A) The FPC implements a $4\times4$ taxel array, with four quadrants beneath each nub-based taxel, addressed as an $8\times8$ row--column matrix.
    During scanning, one column is driven at a time, while one row is connected to the virtual-ground front end.
    Compression reduces the resistance of the silver-fiber resistive layer, increasing current between paired electrodes within a quadrant.
    (B) The assembled sensor stacks the cast silicone and resistive layer over the FPC.
    Normal force raises the four quadrant readings uniformly, whereas shear creates across-quadrant difference.}
    \label{fig:sensor_design}
\end{figure*}

After Section~\ref{sec:background} outlining related work, Section~\ref{sec:design} details the electronic design, the nub-based transduction mechanism validated via finite element analysis, as well as the fabrication process.
% uses finite element analysis to validate our passive nub-based transduction mechanism and predict the effect of nub height. 
Section~\ref{sec:characterization} experimentally evaluates various nub heights and quantifies their force-reconstruction accuracy on physical sensors, resulting in a design with  $0.39~\mathrm{N}$ normal RMSE and $0.16~\mathrm{N}$ shear RMSE with a range of $1$--$5~\mathrm{N}$ normal force and up to $1.5~\mathrm{N}$ shear per taxel.  
% We find that taller nubs improve shear-force reconstruction, with the $3.0~\mathrm{mm}$ design performing best and achieving $0.39~\mathrm{N}$ normal RMSE and $0.16~\mathrm{N}$ shear RMSE on held-out single-taxel contacts over $1$--$5~\mathrm{N}$ normal force and up to $1.5~\mathrm{N}$ shear.
We further assess performance differences between flat, convex, and concave configurations and demonstrate a practical calibration process through short manual interactions against an F/T reference.

Finally, we demonstrate the practical utility of VecTile on compliant gripper fingers in two downstream tasks (Section~\ref{sec:demonstration}). In the scalpel--balloon interaction, we show that the sensor's shear force sensitivity allows a sharp scalpel to push against a fully blown balloon without rupturing it. A visuotactile plug-insertion task (Fig.~\ref{fig:teaser}) shows the sensor's practicality via a learned manipulation policy under partially visual occlusion.

% The contributions of this paper are: (1) a passive nub-based transduction mechanism that converts three-axis contact forces into common and differential pressure responses on a normal-sensitive piezoresistive array; (2) VecTile, a low-cost, flexible $4\times4$ tactile array with local 3-DoF force reconstruction, together with characterization of nub-height effects, curved-surface transfer, practical calibration, and distributed contact; and (3) system-level validation on compliant gripper fingers through a shear-triggered protective reflex and visuotactile plug insertion under visual occlusion.

\section{Related Work}
\label{sec:background}

Existing tactile sensors obtain multi-axis contact information using various forms of transduction hardware. Vision-based sensors such as GelSight \cite{yuan2017gelsight} and DIGIT \cite{lambeta2020digit} use cameras to image the deformation of an elastomer surface, from which recent learning-based methods estimate a resultant 3-axis force \cite{higuera2024sparsh}. However, their cameras, illumination, and optical paths require rigid housings, resulting in relatively bulky units with a limited sensing area. Magnetic sensors instead infer deformation from relative motion between magnets and magnetometers. eFlesh \cite{pattabiraman2025eflesh} and ReSkin \cite{bhirangi2021reskin} place magnetic skins over magnetometer boards to improve replaceability and customizability. However, the active electronics beneath the sensing surface complicate large-area integration and limit how freely the sensor can bend.

Thin capacitive and piezoresistive matrices can form conformable tactile skins, but simple implementations primarily measure normal pressure \cite{wistreich2025dexskin,huang2026flexitac}. CoinFT recovers a six-axis wrench from capacitive layer displacement, but remains a rigid, single sensing unit rather than a distributed skin \cite{choi2025coinft}. 

Our VecTile design utilizes raised nubs as an alternative by mechanically converting shear into differential normal pressure. Prior works have also investigated this mechanism: Huh et al. created a rigid capacitive sensor for slip detection \cite{huh2020dynamically} and Yu et al. proposed a flexible piezoelectric nub array wrapped around finger sections of rubber gloves \cite{yu2016flexible}. While both provide evidence for this mechanism's high performance, neither has created a design that is particularly practical or cost effective in terms of in-house fabrication: the former requires spin coating of silicone on PCB, while the latter requires material coating, etching, and bonding.

% VecTile integrates a passive nub layer with a shared row/column piezoresistive matrix, providing local 3-DoF force sensing without individually wired cells or per-taxel multi-axis electronics. This integration reduces hardware complexity. We further characterize the mechanism through FEA, nub-height experiments, and F/T evaluation across the array on flat and curved mounts (Section~\ref{sec:characterization}).

\begin{figure*}[!b]
    \centering
    \includegraphics[width=0.9\linewidth]{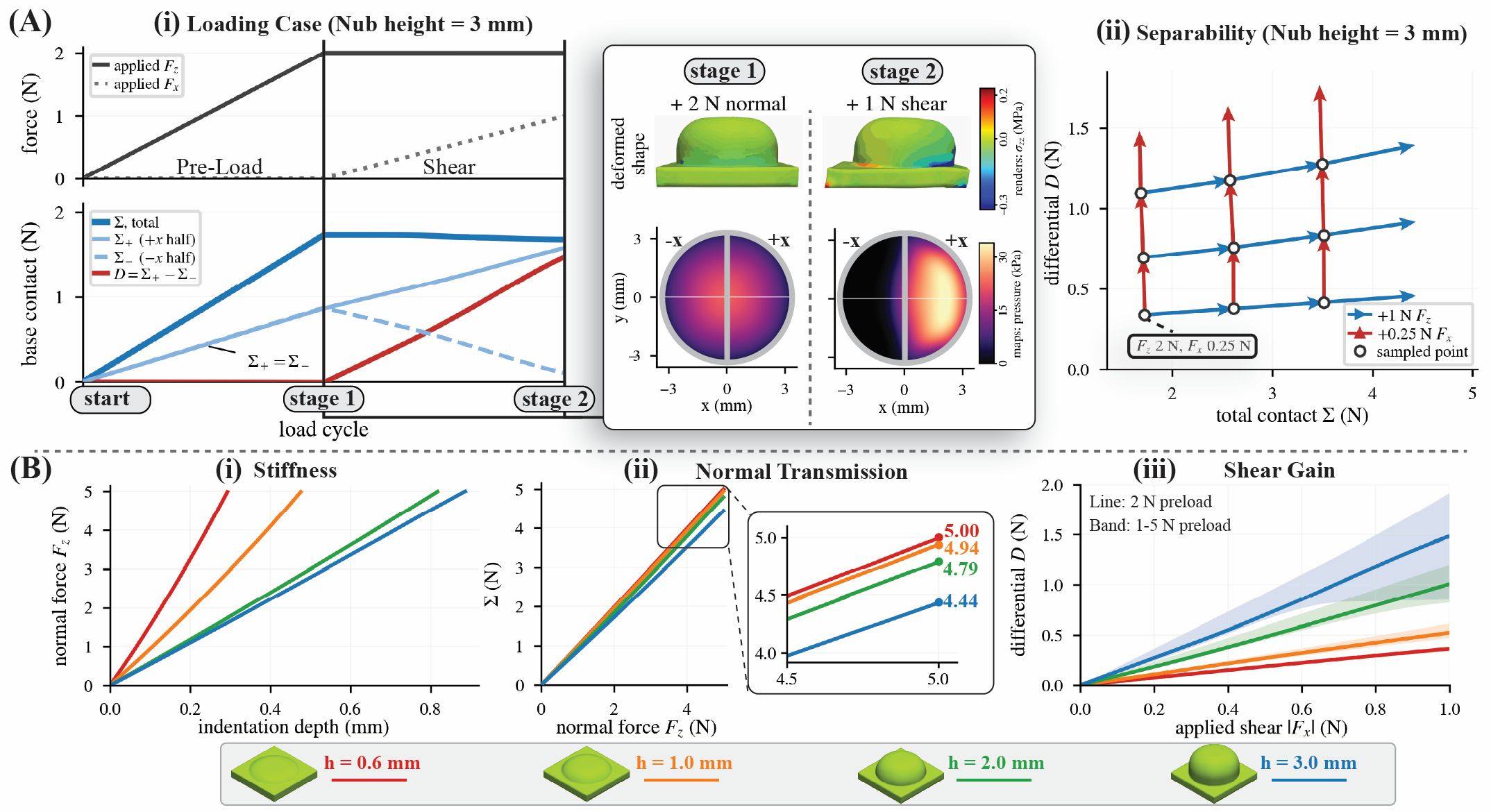}
    \caption{Finite element analysis of the nub-lever mechanism.
    (A) Force encoding and mechanical separability for a $3.0~\mathrm{mm}$ nub.
    (i) Under a $2~\mathrm{N}$ normal preload, the $+x$ and $-x$ half-plane forces are equal.
    Applying $1~\mathrm{N}$ of shear shifts the contact pressure toward the leading half, producing a differential $D$.
    % (ii) Mapping the simulated loads into the $(\Sigma,D)$ reading space shows that normal-force increments act primarily along $\Sigma$, whereas shear-force increments act primarily along $D$.
    % The nonfolding sampled grid indicates local mechanical separability between normal and shear loading over the simulated positive-$x$ envelope.
    (ii) Simulated loads mapped into the $(\Sigma,D)$ reading space. Normal-force increments (blue) move mainly along $\Sigma$, whereas shear-force increments (red) move mainly along $D$. Each $(\Sigma,D)$ reading maps to a unique $(F_z,F_x)$ pair over the simulated positive-$x$ range.
    (B) Effect of nub height on the mechanical response.
    (i) Normal force against indentation depth.
    (ii) Total contact force $\Sigma$ within the nub footprint against applied normal force $F_z$.
    (iii) Differential $D$ against applied shear shows the increase in shear gain with nub height; the lines show the $2~\mathrm{N}$ preload, and the shaded bands show $D$'s span across the $1$--$5~\mathrm{N}$ preload range.}
    \label{fig:nub_simulation}
\end{figure*}

\section{Sensor Design}
\label{sec:design}

\subsection{Sensing Element and Readout Electronics}

Fig.~\ref{fig:sensor_design}(A) presents the sensing element, matrix architecture, and readout electronics.
VecTile is a $4\times4$ taxel array with per-taxel three-axis force sensing, implemented on an FPC.

Each round taxel comprises four piezoresistive quadrants (circular sectors) in a $2\times2$ arrangement. Within the quadrants sit electrode pairs: the outer ones are connected in rows (red), while the inner ones are connected in columns (blue), creating an addressable $8\times8$ row--column matrix. The sensing mechanism relies on variable resistance: a pressure-sensitive silver-fiber fabric (TEK Silver Fiber model 75D, chosen for its local compliance to minimize cross-quadrant influence) is put on top of the FPC; compression lowers its resistance across each electrode pair. We read the 64 quadrants via the Arduino Nano ADC with the driven-column, sensed-row matrix scan illustrated in Fig.~\ref{fig:sensor_design}(A).

% The sensor therefore contains 64 quadrants, which the electronics address as an $8\times8$ row--column matrix.
% Each quadrant contains a electrode pair: one electrode connects to a row and the other connects to a column.

A common flexible resistive architecture~\cite{meribout2024tactile} places the sensing material between two patterned electrode sheets, with the current flowing perpendicular through the pressure-sensitive layer.
VecTile instead places both electrodes of every quadrant on one single FPC, relying on lateral current flow (see ``Cross Section View'' in Fig.~\ref{fig:sensor_design}(A)). On top of a simpler assembly, the main advantage is in the perfect electrode alignment without the need for precision layer positioning.

Initially, we followed the simple-voltage-divider readout approach from \cite{huang2026flexitac}, but experienced early saturation due to additional current paths when multiple nubs are loaded simultaneously. This issue was solved via an additional LM358 transimpedance amplifier, holding the non-activated columns at a $0~\mathrm{V}$ virtual ground. A full list of electronic components can be found in the figure.

For a selected quadrant $(i,j)$, the ideal output is given by relation $ V_{\mathrm{ADC},ij}    \simeq V_{\mathrm{drive}}R_{\mathrm{ref}}/R_{ij}$,
% \begin{equation}
%     V_{\mathrm{ADC},ij}
%     \simeq V_{\mathrm{drive}}\frac{R_{\mathrm{ref}}}{R_{ij}},
%     \label{eq:quadrant_readout}
% \end{equation}
where $R_{ij}$ is the pressure-dependent resistance of quadrant $(i,j)$, $V_{\mathrm{drive}}$ is the selected-column voltage, and $R_{\mathrm{ref}}$ is the transimpedance feedback resistance.
Compression lowers $R_{ij}$ and therefore increases the ADC output.
We implement $R_{\mathrm{ref}}$ with a potentiometer, allowing the readout gain to accommodate resistive layers with different working resistances.
For the present build, $R_{\mathrm{ref}}=10~\mathrm{k}\Omega$; the fixed resistors in the two-stage front end are also $10~\mathrm{k}\Omega$.

The Arduino ADC returns a 10-bit conversion for each quadrant, after which the firmware discards the two least-significant bits and transmits one 8-bit value per quadrant.
Each 64-byte frame payload is transmitted at $2~\mathrm{Mbaud}$ and a measured rate of approximately $343~\mathrm{Hz}$.

\subsection{Nub-Lever Mechanism}
\label{sec:nub_lever}

\subsubsection{Working Principle}
\label{sec:nub_working_principle}

Fig.~\ref{fig:sensor_design}(B) shows the assembled sensor stack and the nub-lever working principle.
Each cast silicone nub covers four sensing quadrants and converts a three-axis contact force into four local normal-force signals.
A normal force $F_z$ compresses the nub and raises the four quadrant readings together.
A shear force $F_x$ or $F_y$ tilts the nub and transfers normal load from the trailing side to the leading side.
The mechanism therefore encodes normal force in the balanced response and shear force in the imbalance between opposing quadrant pairs.

\subsubsection{Finite Element Model}

To quantify nub-induced pressure redistribution independently of piezoresistive transduction and electrical readout, we developed a finite element model of a single nub under combined normal and shear loading (Fig.~\ref{fig:nub_simulation}).
The simulation assumes nominal in-plane symmetry and therefore considers only the positive $F_x$ shear force case. 
We selected $3.0~\mathrm{mm}$ as the maximum nub height to retain a thin sensor profile and simulated heights of $0.6$, $1.0$, $2.0$, and $3.0~\mathrm{mm}$ to quantify the effect of reducing nub height.
The model uses Dragon Skin 30 and the fabricated nub geometry (near-incompressible Mooney--Rivlin model~\cite{mooney1940theory}, $C_{10}=0.198~\mathrm{MPa}$ and $C_{01}=0.049~\mathrm{MPa}$; four $1\times1~\mathrm{mm}$ fixed corner patches on the cap underside; remaining underside in frictionless rigid plane; SOLID187 elements with mesh size $\min(0.4~\mathrm{mm},h/2)$).
For each of the four nub heights, we ramp $F_z$ to a preload between $1$ and $5~\mathrm{N}$ and then ramp $F_x$ to $1~\mathrm{N}$ while holding $F_z$, giving 20 loading cases.

We split the nub footprint along the $x$-axis and integrate the normal contact force over its $+x$ and $-x$ halves, denoted $\Sigma_+$ and $\Sigma_-$, respectively.
Their sum gives the total contact reading, while their difference measures the contact imbalance:
$\Sigma=\Sigma_{+}+\Sigma_{-}$ and $D=\Sigma_{+}-\Sigma_{-}$, 
% \begin{equation}
%     \Sigma = \Sigma_+ + \Sigma_-,
%     \qquad
%     D = \Sigma_+ - \Sigma_-.
%     \label{eq:sim_channels}
% \end{equation}
where $\Sigma$ is the total normal contact force transmitted within the nub footprint, and $D$ (the differential force) increases as contact moves from the $-x$ half to the $+x$ half.
We solved the model using large-deflection statics in ANSYS MAPDL 2024 R2.

\subsubsection{Mechanical Separability of Forces}

Fig.~\ref{fig:nub_simulation}(A.i) illustrates the forward encoding from the applied forces $(F_z,F_x)$ to the base readings $(\Sigma,D)$.
Under a $2~\mathrm{N}$ normal load, the $+x$ and $-x$ halves carry approximately equal contact force, giving $D\approx0$ and $\Sigma=1.73~\mathrm{N}$.
Applying $1~\mathrm{N}$ of shear at the same normal load shifts contact from the $-x$ half to the $+x$ half, increasing $D$ to $1.47~\mathrm{N}$ while changing $\Sigma$ only slightly.
% , from $1.73$ to $1.68~\mathrm{N}$.
Normal force therefore acts mainly through the total contact reading $\Sigma$, whereas shear force acts mainly through the contact imbalance $D$.
The difference between $\Sigma$ and the applied $F_z$ is load carried outside the measured nub footprint.

To evaluate whether normal and shear forces remain distinguishable, Fig.~\ref{fig:nub_simulation}(A.ii) shows the change in $(\Sigma,D)$ from multiple preload and shear states.
From each state, blue arrows show the response to adding $1~\mathrm{N}$ of $F_z$ at fixed $F_x$, while red arrows show the response to adding $0.25~\mathrm{N}$ of $F_x$ at fixed $F_z$.
Across all tested states, normal-force increments act mainly through $\Sigma$, whereas shear-force increments act mainly through $D$.
The two response directions remain separated by $86$--$90^\circ$ and retain a consistent orientation across the grid.
This indicates local mechanical separability and supports inversion from $(\Sigma,D)$ to $(F_z,F_x)$ within the sampled positive-$x$ range.

\subsubsection{Simulated Effect of Nub Height}
\label{sec:sim_height}

Because nub height changes both the compliance and lever arm, we compare the four heights under the same loading conditions in Fig.~\ref{fig:nub_simulation}(B).
Fig.~\ref{fig:nub_simulation}(B.i) shows that taller nubs require more indentation to support the same normal force and are therefore more compliant.
Despite this change in stiffness, Fig.~\ref{fig:nub_simulation}(B.ii) shows that most of the applied normal force reaches the nub footprint for every height.
At $F_z=5~\mathrm{N}$, $\Sigma$ decreases from $5.00~\mathrm{N}$ for the $0.6~\mathrm{mm}$ nub to $4.44~\mathrm{N}$ for the $3.0~\mathrm{mm}$ nub, while the remaining load is carried outside the footprint.
In contrast, Fig.~\ref{fig:nub_simulation}(B.iii) shows a strong increase in the differential response with nub height.
The simulation therefore predicts that nub height primarily controls shear gain while preserving most of the normal force transmitted to the taxel footprint.
This result motivates the experimental height sweep in Section~\ref{sec:force_accuracy}.

\subsection{Sensor Fabrication}
\label{sec:fabrication}

\begin{figure}[t]
    \centering
    \includegraphics[width=1\linewidth]{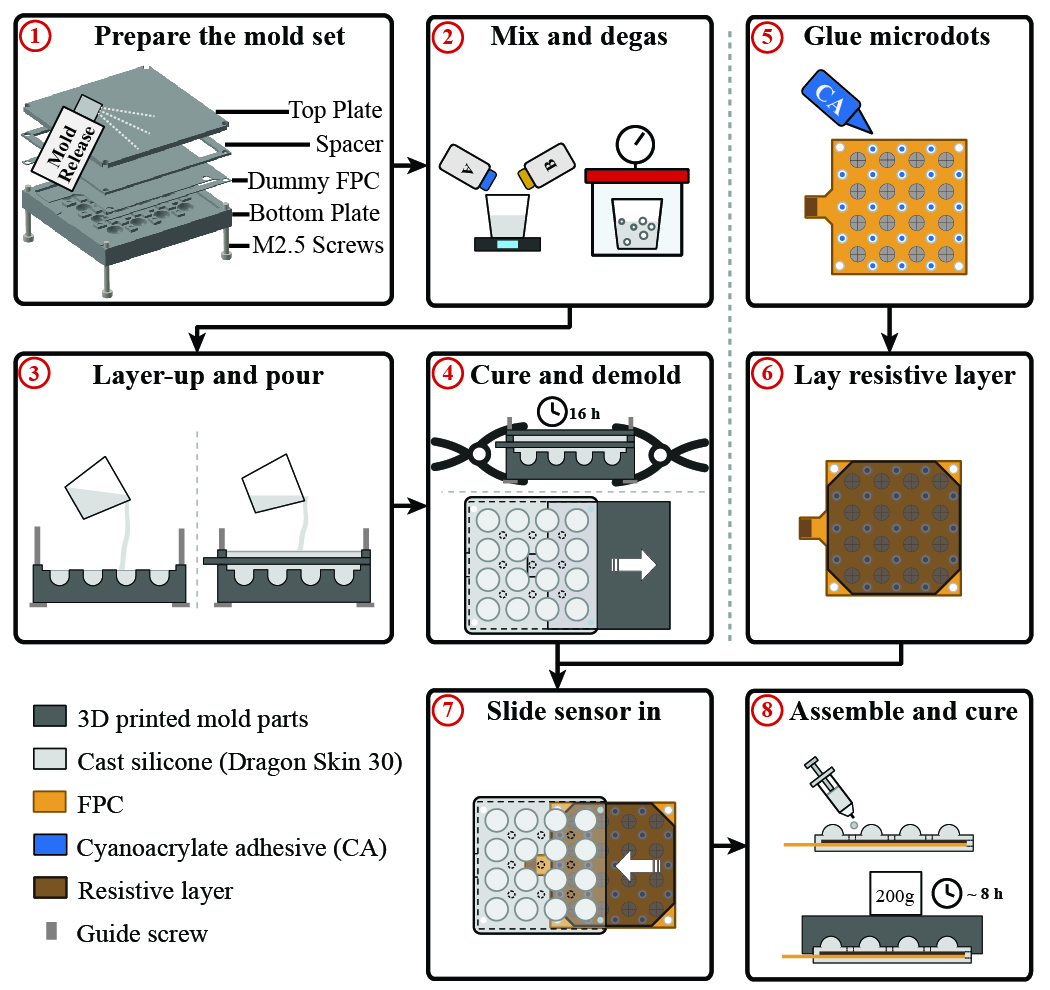}
    \caption{VecTile fabrication.
    (1) Mold release is applied to the 3D-printed PLA mold components.
    (2) Dragon Skin 30 is mixed and vacuum-degassed.
    (3) The bottom plate is filled with the degassed silicone, after which the dummy FPC and spacer are positioned and a second pour completes the mold stack.
    (4) The mold is closed, clamped, and cured for $16~\mathrm{h}$; the silicone layer is then de-molded and the dummy FPC is slid out.
    (5) Cyanoacrylate glue dots are applied only at the marked locations on the FPC.
    (6) The resistive layer is placed over the FPC.
    (7) The FPC and resistive layer are slid into the molded silicone layer produced in step~(4).
    (8) A small batch of Dragon Skin 30, prepared as in step~(2), is injected through the openings; the bottom plate from step~(1) is reused as a press under a $200~\mathrm{g}$ load during the approximately $8~\mathrm{h}$ cure.}
    \label{fig:sensor_fabrication}
\end{figure}

The fabrication process involves only Smooth-On Dragon Skin 30 silicone molding, bonding via commercially available glue, and simple assembly. No equipment is needed other than a 3D printer for molds (we used a Bambu Lab P1S for PLA molds), a degassing chamber (in our case a generic vacuum chamber under 150 USD), and readily available tools such as a syringe, plastic clamps, and a lab scale. Fig.~\ref{fig:sensor_fabrication} shows the complete procedure.

% must align each nub with its four quadrants and keep the active electrode interfaces mechanically free. 

During molding, a 3D printed $0.4~\mathrm{mm}$-thick PLA dummy FPC approximates the combined thickness of the sensor FPC and resistive layer and reserves a matching insertion pocket.
The real sensing stack then slides into this pocket, so the mold geometry aligns the FPC, resistive layer, and silicone.
% The mold components were printed in PLA on a Bambu Lab P1S.

Two bonding materials are used to secure the stack while preserving its local compliance.
Because silicone adheres poorly to the FPC, cyanoacrylate glue dots bond the resistive layer to the FPC at the marked positions between sensing regions.
This glue is not used between the resistive layer and molded silicone layer because it locally hardens the silicone and constrains its deformation.
Instead, a small batch of Dragon Skin 30 is injected through the mold openings to bond the resistive layer to the silicone layer.
Both bonding materials remain outside the electrode areas, leaving the sensing interfaces locally compliant for pressure redistribution.
Because the process adds no rigid backing and confines bonding to small regions, the assembled sensing tile remains flexible.

% The same workflow produces different nub heights by changing the nub-cavity geometry while leaving the remaining fabrication and assembly steps unchanged.
% We used this method to fabricate all nub-height variants evaluated in the later sections.

At the stated order quantities, the sensing layer plus readout costs 34.67 USD per unit, but can be reduced to under 15 USD if a cheaper microcontroller is used (Table~\ref{tab:cost}).

\begin{table}[t]
\centering
\caption{Bill of Materials}
\scriptsize
\renewcommand{\arraystretch}{0.96}
\setlength{\tabcolsep}{6pt}
\begin{tabular}{ll c}
\toprule
 & \textbf{Item} & \textbf{Cost (USD)} \\
\midrule
\multirow{6}{*}{\textbf{Sensor}}
 & FPC (30-pc order) & 0.89 \\
 & Resistive layer ($\approx$ 0.004 m$^2$ at \$30/m$^2$) & 0.12 \\
 & Dragon Skin 30 ($\approx$ 15\,g at \$20/lb trial kit) & 0.70 \\
 & FFC cable + connector set & 1.50 \\
 & Mold filament ($\approx$ 30\,g at \$12/kg) & 0.36 \\
 \cmidrule(lr){2-3}
 & Subtotal & 3.57 \\
\midrule
\multirow{3}{*}{\textbf{Readout}}
 & PCB, assembled (10-pc order) & 6.10 \\
 & Arduino Nano\textsuperscript{a}  & 25.00 \\
 \cmidrule(lr){2-3}
 & Subtotal & 31.10 \\
\midrule
 & \textbf{Total per unit} & \textbf{34.67} \\
\bottomrule
\end{tabular}
\label{tab:cost}
\vspace{1mm}

\parbox{\linewidth}{\scriptsize \textsuperscript{a}~A Nano-compatible ATmega328P development board costs approximately \$5, reducing the projected total to \$14.67.}
\end{table}

\section{Sensor Characterization}
\label{sec:characterization}

We characterize the physical sensor performance, and test whether the mechanically separable response predicted by the
simulation remains observable after resistive transduction and electronic
readout. We first characterize the resistive layer, then evaluate how nub
height and curved mounting affect the raw response and decoded three-axis
accuracy.

\subsection{Transduction of the Resistive Layer}
\label{sec:layer_transduction}

\begin{figure}[!b]
    \centering
    \includegraphics[width=\linewidth]{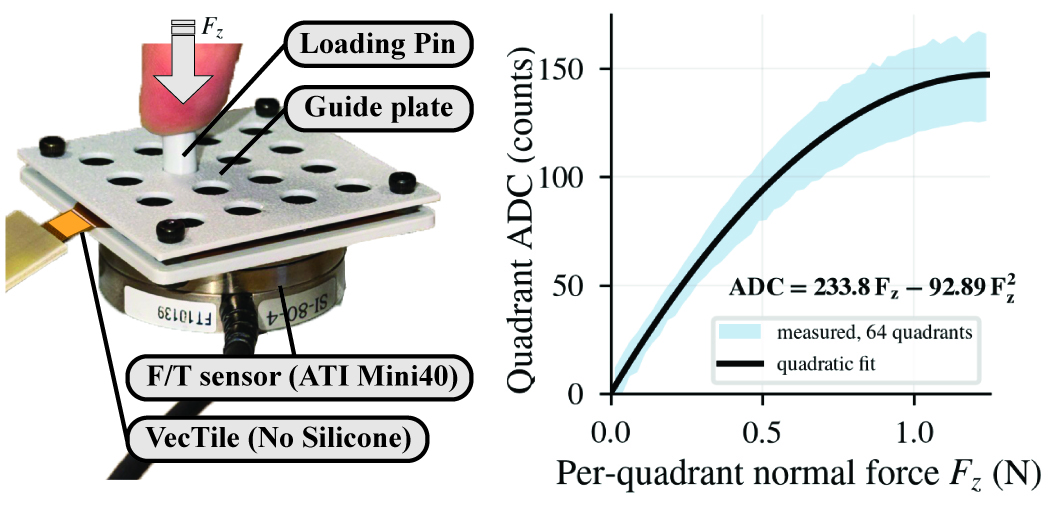}
    \caption{Left: VecTile without the silicone layer mounted on an ATI Mini40 force/torque sensor, with a guide plate aligning the loading pin to each taxel. Right: Per-quadrant ADC response to normal force; the shaded envelope spans 64 quadrants. The quadratic is fitted over $F_z \leq 1.25~\mathrm{N}$.}
    \label{fig:resistive_transduction}
\end{figure}

We first characterized the resistive layer separately. As shown in Fig.~\ref{fig:resistive_transduction}, the FPC and resistive fabric were mounted directly on an ATI Mini40 F/T sensor without the silicone layer. A guide plate aligned the loading pin with each taxel. The pin covered the four quadrants of one taxel, so the applied normal force and summed counts were divided by four to report the response on a per-quadrant basis. We applied 20 manual loading cycles to each of the 16 taxels.

Fig.~\ref{fig:resistive_transduction} shows that the ADC response becomes progressively less sensitive as the quadrant force approaches $1.25~\mathrm{N}$. This is not a hard force limit, but additional force above this region produces little change in the reading and is therefore more difficult to resolve accurately. Under the four-quadrant normalization used in the plot, $1.25~\mathrm{N}$ per quadrant corresponds to approximately $5~\mathrm{N}$ per taxel. We therefore selected $1$--$5~\mathrm{N}$ as the normal-force range for each nub for the characterization of the whole sensor. 

\begin{figure}[t]
    \centering
    \includegraphics[width=\linewidth]{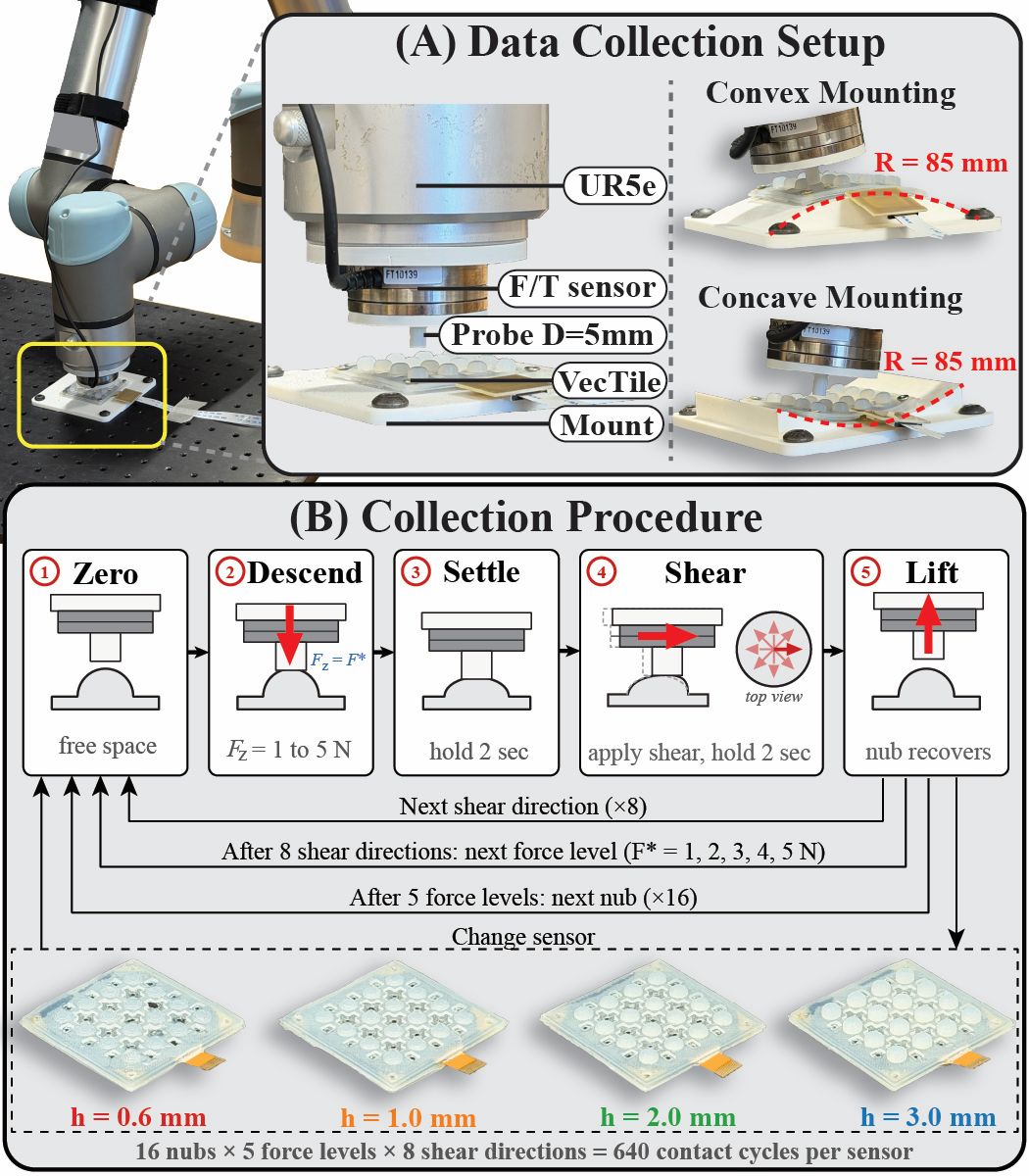}
    \caption{Sensor characterization. (A) Experimental setup for flat and
    $R=85~\mathrm{mm}$ curved mountings. (B) Data collection procedure.}
    \label{fig:characterization_setup}
\end{figure}

\subsection{Data Collection Setup}
\label{sec:characterization_setup}

To compare nub heights under consistent loading conditions, we used the UR5e robot arm setup in Fig.~\ref{fig:characterization_setup} to perform controlled normal and tangential interactions. An ATI Mini40 F/T sensor between the rigid probe and the UR5e wrist collects ground truth force data.
% A rigid probe and an ATI Mini40 F/T sensor were mounted on the robot, allowing the same force levels and shear headings to be applied to every nub while recording the reference wrench.
We tested four flat sensor builds with nub heights $h=0.6$, $1.0$, $2.0$, and $3.0~\mathrm{mm}$.
The $3.0~\mathrm{mm}$ geometry was additionally tested on convex and concave mounts with radii $R=85~\mathrm{mm}$.
% ; for these tests, the reference wrench was expressed in the local surface frame.

The procedure starts with a non-contact baseline used to zero both VecTile and the F/T sensor, followed by the probe perpendicularly pressing into a nub to a target preload force, holding for 2 seconds, then moving laterally by 2 mm under repeatable displacement control, holding for another 2 seconds, then releasing. No slipping was observed during this process. Repeating this for all 5 target normal preload forces $F_z^*\in\{1,2,3,4,5\}~\mathrm{N}$, eight uniformly distributed shear directions spaced 45$^\circ$  apart, and 16 nubs, we collected 640 interaction sequences per sensor. With data saved at 86.5 Hz, 348,800 data points were collected.

% Each trial began with a no-contact baseline. The probe then descended to a target normal preload $F_z^*\in\{1,2,3,4,5\}~\mathrm{N}$ and held its position for $2~\mathrm{s}$. It subsequently moved $2~\mathrm{mm}$ along one of eight shear directions, held for $2~\mathrm{s}$, and returned. The shear motion was displacement-controlled, while the resulting force, reaching up to approximately $1.5~\mathrm{N}$. Synchronized Mini40 measurements provided the reference force labels. This procedure yielded 640 contacts per sensor.

\subsection{Force Reconstruction Across Nub Heights}
\label{sec:force_accuracy}

\begin{figure}[t]
    \centering
    \includegraphics[width=\linewidth]{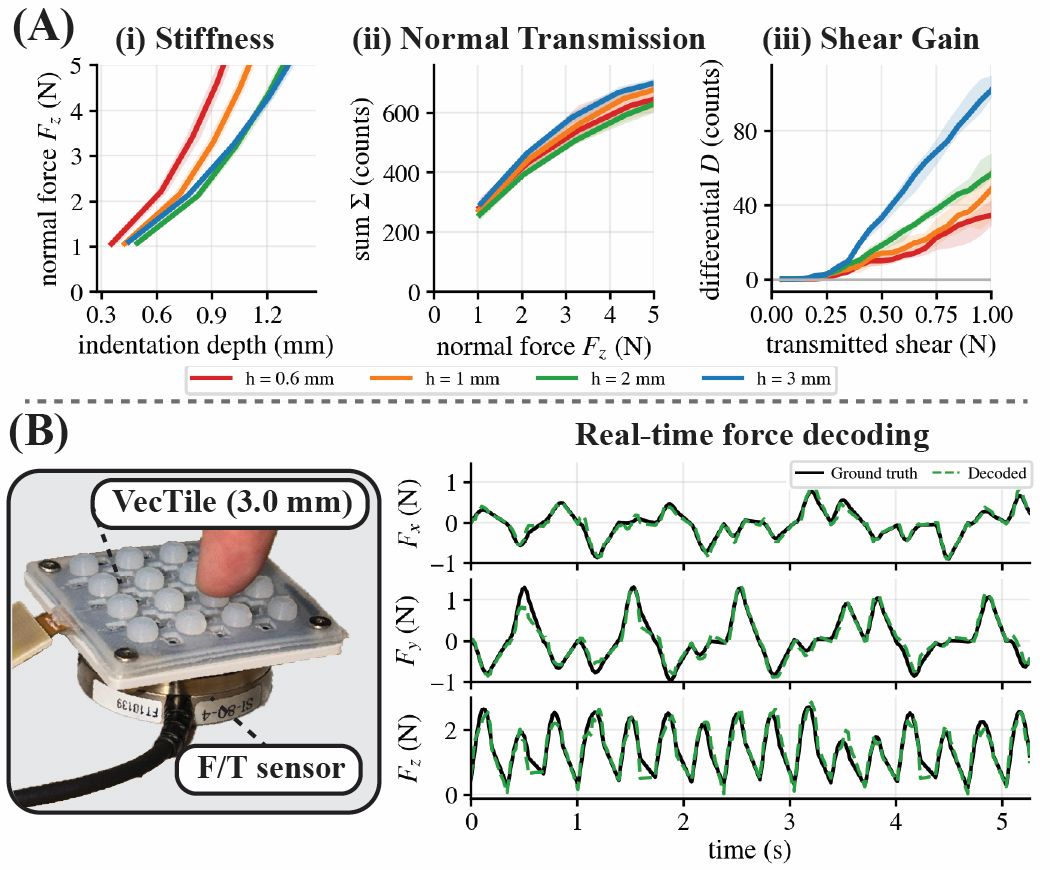}
    \caption{Measured response and manual force decoding. (A) Effect of nub height on (i) stiffness, (ii) the normal-force feature $\Sigma$, and (iii) the shear-force feature $D$. Solid lines are the median while the bands show the range, both across 16 taxels. (B) Manual bench validation of a $3.0~\mathrm{mm}$ VecTile mounted on an ATI Mini40: hand loading (left) and a real-time comparison between the VecTile estimates and F/T reference measurements (right).}
    \label{fig:raw_height_response}
\end{figure}

% We examine the raw readings to determine whether the effects of nub height predicted by the simulation remain visible after piezoresistive transduction.
% For each taxel, we sum the four quadrant readings to form the normal feature $\Sigma$ and project the difference between opposing quadrants onto the commanded shear heading to form the shear feature $D$, correspond to the simulation.
% The lines show the median over 16 taxels, and the bands show the interquartile range.
We examine whether the effects of nub height predicted by simulation remain visible after piezoresistive transduction, shown in Fig.~\ref{fig:raw_height_response}(A). For each taxel, the quadrant sum defines the normal feature $\Sigma$. The opposing-quadrant differences $D_x$ and $D_y$ are projected onto the shear direction $\theta$ as $D=-(D_x\cos\theta+D_y\sin\theta)$. 
% The lines and bands show the median and interquartile range across 16 taxels.

% Fig.~\ref{fig:raw_height_response}(A.i) first shows that reaching the same normal force requires more indentation as nub height increases.
% This result confirms that taller nubs are more compliant, matching the simulated stiffness trend.

% Despite this change in compliance, Fig.~\ref{fig:raw_height_response}(A.ii) shows that $\Sigma$ increases similarly with normal force for all four heights.
% The overlapping curves show that nub height has little effect on the normal ADC response of the assembled sensor.

% The main effect of nub height appears in the shear channel in Fig.~\ref{fig:raw_height_response}(A.iii), where the curves separate further as the shear force increases.

The results from the physical testing agree with the FEA results from Fig.~\ref{fig:nub_simulation}(B): despite a larger nub height resulting in lower overall stiffness, there is little effect on the actual ADC response in terms of normal force; shear sensitivity significantly improves with larger nub height. At $1~\mathrm{N}$ of transmitted shear, the median projected differential increases consistently with nub height, from 34.2 counts for the $0.6~\mathrm{mm}$ nub to 101.6 counts for the $3.0~\mathrm{mm}$ nub.
% The results show that taller nubs increase compliance and shear sensitivity while leaving the normal response largely unchanged, consistent with simulation.

For each taxel, we fit a quadratic decoder that maps its four baseline-subtracted quadrant readings to each force component:
\begin{equation*}
\hat{F}_a=c_a+\sum_{i=1}^{4}w_{a,i}q_i
+\sum_{1\leq i\leq j\leq4}w_{a,ij}q_iq_j,
\quad a\in\{x,y,z\},
\end{equation*}

where $\hat{F}_a$ is the estimated force component, $q_i$ is the ADC reading of quadrant $i$, and $c_a$, $w_{a,i}$, and $w_{a,ij}$ are coefficients fitted by ordinary least squares.

We evaluated each fitted map using a four-fold direction-rotation split: at each normal-force level, one opposing pair of shear directions was withheld, and the withheld pair was rotated across force levels and folds.
% Each force--heading combination was therefore evaluated once, while every heading remained represented in the training data at other force levels.
Table~\ref{tab:accuracy} reports normal~$(F_z)$ and shear RMSEs of $(F_x,F_y)$ on the held-out contacts.
The 4 nub-height designs achieve similar normal RMSE, while the $3.0~\mathrm{mm}$ design achieves $0.16~\mathrm{N}$ shear RMSE compared with $0.29~\mathrm{N}$ for the $0.6~\mathrm{mm}$ design.
The wider shear count range of the taller nub improves force reconstruction without sacrificing normal-force accuracy.

To investigate how a sensor calibrated in a flat configuration behaves when bent, we mounted the 3.0 mm sensor on the curved fittings ($R=85~\mathrm{mm}$). The quadratic decoder was directly taken from the previous investigation (fitted in a flat configuration). Running the aforementioned procedure in this curved configuration, the resulting sensor predictions are directly compared with the raw F/T sensor readings. Relative to flat mounting, the curved mounts increased normal and shear RMSE by an average of $0.05~\mathrm{N}$ and $0.06~\mathrm{N}$ respectively.    

% The $3.0~\mathrm{mm}$ design provides the lowest RMSE, so we use it to evaluate curvature transfer.
% We mounted the same sensor on the $R=85~\mathrm{mm}$ convex and concave fixtures in Fig.~\ref{fig:characterization_setup}(A) and repeated the controlled loading sequence, with the probe approaching along the local surface normal before applying shear.
% We applied the quadratic model previously fitted on the flat mount directly to the raw curved-mount readings, and compared its predictions with the F/T reference in the local tangent frame.
% Relative to flat mounting, the curved mounts increased normal and shear RMSE by an average of $0.05~\mathrm{N}$ and $0.06~\mathrm{N}$ respectively, without retraining.

\begin{table}[t]
% \centering
% \caption{Force RMSE across flat nub heights and curved mounts.}
% \scriptsize
% \renewcommand{\arraystretch}{1.05}
% \setlength{\tabcolsep}{4.5pt}
% \begin{tabular}{l cccc cc}
% \toprule
% \textbf{RMSE (N)} & \multicolumn{4}{c}{\textbf{Flat: nub height (mm)}} & \multicolumn{2}{c}{\textbf{Curved: h = 3.0 mm}} \\
% \cmidrule(lr){2-5}\cmidrule(lr){6-7}
%  & 0.6 & 1.0 & 2.0 & 3.0 & convex & concave \\
% \midrule
% $F_z$ & 0.45 & 0.45 & 0.41 & \textbf{0.39} & 0.45 & 0.42 \\
% Shear & 0.29 & 0.28 & 0.24 & \textbf{0.16} & 0.21 & 0.23 \\
% \bottomrule
% \end{tabular}
% \vspace{1mm}
\centering
\caption{Force RMSE across nub heights and curved mounts.}
\scriptsize
\renewcommand{\arraystretch}{1.05}
\setlength{\tabcolsep}{3pt}
\resizebox{\linewidth}{!}{%
\begin{tabular}{l cccc cc}
\toprule
\textbf{RMSE (N)} & \multicolumn{4}{c}{\textbf{Nub height (mm)}} & \multicolumn{2}{c}{\textbf{Curved, 3.0 mm}} \\
\cmidrule(lr){2-5}\cmidrule(lr){6-7}
 & 0.6 & 1.0 & 2.0 & 3.0 & convex & concave \\
\midrule
$F_z$ & 0.45\,$\pm$\,0.06 & 0.45\,$\pm$\,0.05 & 0.41\,$\pm$\,0.04 & \textbf{0.39\,$\pm$\,0.03} & 0.45\,$\pm$\,0.06 & 0.42\,$\pm$\,0.04 \\
Shear & 0.29\,$\pm$\,0.04 & 0.28\,$\pm$\,0.04 & 0.24\,$\pm$\,0.03 & \textbf{0.16\,$\pm$\,0.02} & 0.21\,$\pm$\,0.04 & 0.23\,$\pm$\,0.03 \\
\bottomrule
\end{tabular}%
}
\vspace{1mm}

{\scriptsize Mean $\pm$ standard deviation of the per-taxel RMSE across the 16 taxels. 

Curved results use the model fitted on the flat $3.0~\mathrm{mm}$ sensor. 

Evaluation range: $F_z\in[1,5]~\mathrm{N}$ and $\|\mathbf{F}_{xy}\|_2\in[0,1.5]~\mathrm{N}$.\par}
\label{tab:accuracy}
\end{table}

\subsection{Practical Calibration \& Evaluation }
\label{sec:practical_calibration}

After selecting the $3.0~\mathrm{mm}$ design, we evaluated whether the sensor could be calibrated using a short manual data collection lasting only about a handful of minutes, and whether this calibration can be useful in less controlled, real-life contact interactions.

We mounted VecTile on the ATI Mini40, as shown in Fig.~\ref{fig:raw_height_response}(B), and performed 10 fast manual loading interactions on each nub, with a median duration of 0.55 s per interaction. During each interaction, the operator first pressed normally, then moved the contact in a random circular trajectory. The decoder was fitted on data collected at 12.5 Hz, leading to around 1,400 data points in total. Then, nubs were randomly pressed and moved again, leading to 240 testing data points. 

The manually calibrated model achieved RMSEs of $0.43~\mathrm{N}$ for normal force and $0.22~\mathrm{N}$ for shear force per taxel. Fig.~\ref{fig:raw_height_response}(B) shows a real-time comparison between the VecTile estimates and the F/T reference. 
% The observed offset between the rising edges of the two signals was approximately $25~\mathrm{ms}$, including sensor response, USB acquisition, and quadratic decoding.
The cross-correlation gave a median VecTile-to-F/T lag of $25~\mathrm{ms}$, including sensor response, USB acquisition, and quadratic decoding.

In addition, we investigated multi-nub force prediction following the manual calibration. The operator randomly pressed down and applied shear forces 65 times with the palm and fingers across portions of the sensor, with 624 frames recorded. A median of 8 nubs were contacted for each press, with 7 of these presses contacting all 16 nubs. Loads up to $15.54~\mathrm{N}$ and $5.83~\mathrm{N}$ were recorded by the F/T sensor. VecTile predictions are simply taken as sums of per-taxel normal and shear forces. A normal RMSE of $0.71~\mathrm{N}$ and a shear RMSE of $0.43~\mathrm{N}$ were achieved.   

% We also evaluated distributed contacts by loading multiple nubs simultaneously. For loads reaching $15.54~\mathrm{N}$ in normal force and $5.83~\mathrm{N}$ in shear, the sum of the per-taxel predictions achieved a normal RMSE of $0.71~\mathrm{N}$ and a shear RMSE of $0.43~\mathrm{N}$.
% The model achieves $0.43~\mathrm{N}$ normal RMSE and $0.22~\mathrm{N}$ shear RMSE. Fig.~\ref{fig:raw_height_response}(B) shows a real-time interaction and decoding sample.  We further evaluated distributed contact by manually pressing and rubbing across the sensor surface. Summing the per-taxel force estimates, VecTile achieved $0.71~\mathrm{N}$ normal-force RMSE and $0.43~\mathrm{N}$ shear-vector RMSE for loads up to $15.6~\mathrm{N}$ normal and $6.1~\mathrm{N}$ shear.

% These results demonstrate a lightweight within-session calibration procedure.

\begin{figure}[b!]
    \centering
    \includegraphics[width=\linewidth]{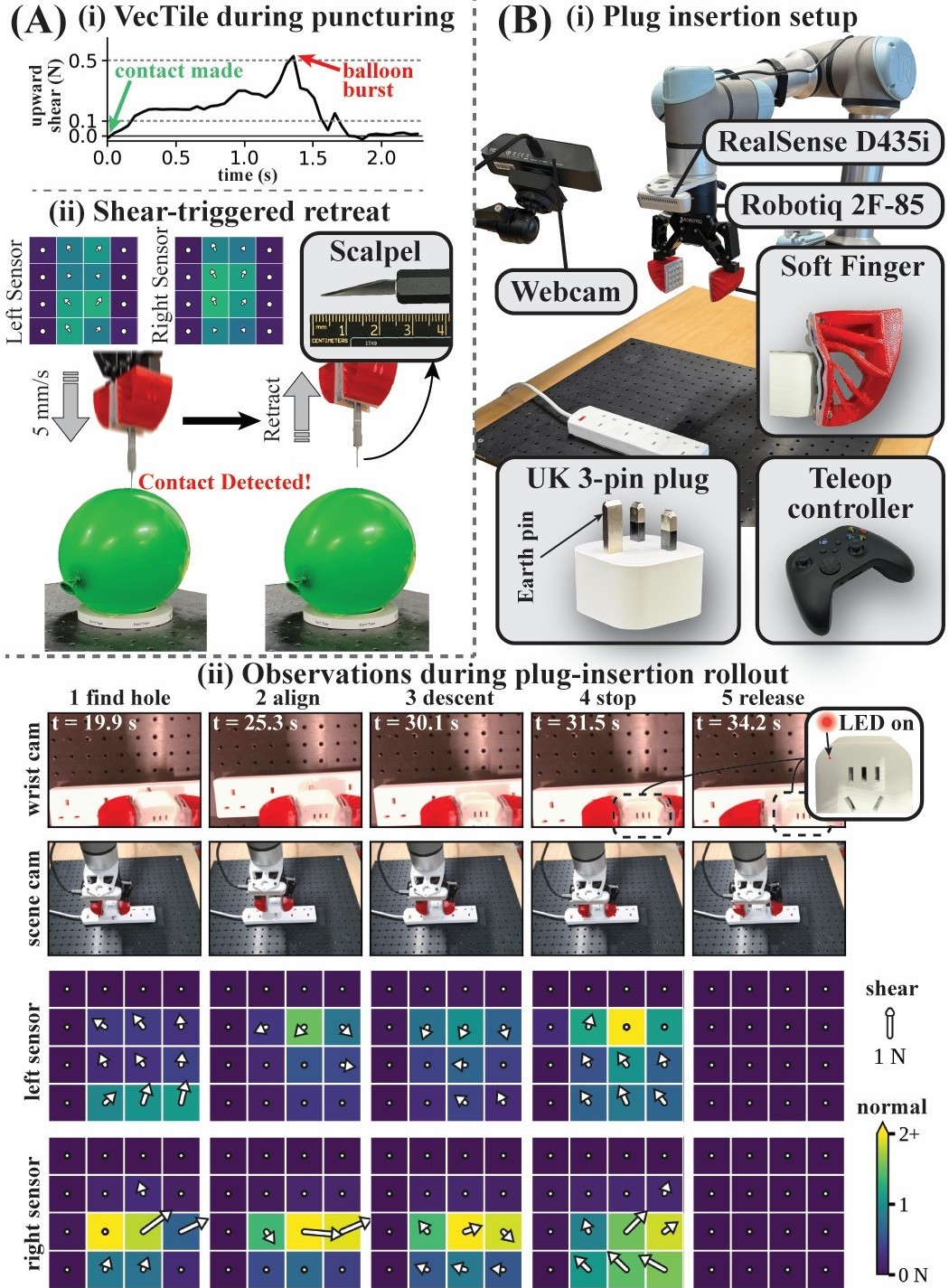}
    \caption{VecTile demonstrations. (A) Scalpel--balloon reflex: (i) shear measured during a burst trial and (ii) shear-triggered retraction before puncture. (B) Visuotactile plug insertion: (i) Hardware setup and (ii) Camera views and fingertip tactile readings during a 3-DoF tactile sensing plug-insertion rollout.}
    \label{fig:demo}
\end{figure}

\section{Sensor Demonstration}
\label{sec:demonstration}
To further demonstrate the practicality of VecTile, we integrated it into two closed-loop robotic manipulation tasks. Task A showcases shear sensitivity by pressing a sharp scalpel against a fully inflated balloon, and retracting once contact is detected without bursting the balloon. Task B evaluates VecTile in a contact-rich plug insertion task using a visuotactile Diffusion Policy \cite{chi2025diffusion}. For both tasks, the sensor used the manual fast-calibration method described in the previous section. Demonstration videos are provided as supplementary material.

We designed a pair of compliant Fin-Ray type fingers for a Robotiq 2F-85 gripper mounted on the UR5e, as shown in Fig.~\ref{fig:demo}. The fingers were printed from generic Amazon Basics 95A TPU using a Prusa MK3S printer. We attached VecTile arrays with $3.0~\mathrm{mm}$ nubs to each finger. Their flexibility conforms to the bendable finger surfaces. 

\subsection{Shear-Aware Scalpel--Balloon Reflex}
In an initial trial that deliberately burst the balloon, the gripper held a scalpel and moved it toward an inflated balloon at $5~\mathrm{mm/s}$. The estimated shear-force change, summed over the 16 taxels of one fingertip along the loading direction, reached approximately $0.5~\mathrm{N}$ before rupture (Fig.~\ref{fig:demo}(A.i)). We then set a per-taxel shear threshold of $0.1~\mathrm{N}$ and conducted the reflex experiment (Fig.~\ref{fig:demo}(A.ii)). The robot arm retracted when the change in shear-force magnitude at any taxel exceeded this threshold. Across 20 repeated trials, scalpel contact triggered retraction before puncture in every instance. In a matched normal-only condition using the same motion and threshold, the normal-force change did not reach the threshold before the balloon burst.

% The gripper held a knife and moved it toward an inflated balloon at $5~\mathrm{mm/s}$ (Fig.~\ref{fig:demo}(B)). Contact with the balloon transmitted a shear load through the knife to the fingertip sensors. The controller commanded retraction when the change in shear-force magnitude at any taxel exceeded $0.1~\mathrm{N}$. Across 20 repeated trials, knife contact triggered retraction before puncture in every instance. In a matched normal-only condition using the same motion and threshold, the normal-force change did not reach the threshold before the balloon burst.

\subsection{Visuotactile Plug Insertion}

We selected plug insertion to evaluate VecTile in contact-rich manipulation, Fig.~\ref{fig:demo}(B.i) shows the setup. The task began with the UK 3-pin plug held by the gripper and required the robot to align and insert it into a socket. A wrist camera (Intel RealSense D435i) and a fixed scene camera (Maxhub UC W20) provided two views. 
% During final alignment, the plug and gripper partially occluded the prongs and socket apertures. VecTile complemented these visual observations by measuring local normal and tangential loading transmitted through the plug during engagement and seating.

We collected 84 demonstrations by teleoperating the robot with an Xbox controller at six socket locations spanning an approximately $100 \times 30~\mathrm{mm}$ region. Each location contributed 7 direct (no corrective adjustment needed) and 7 corrective (direct insertion failed, followed by small adjustments) insertion trajectories.

Independent ResNet-18 \cite{he2016deep} encoders mapped each RGB view to a 512-dimensional feature. Each tactile observation comprised 96 baseline-subtracted channels ($2$ fingers $\times$ $16$ taxels $\times$ $3$ axes), which we projected onto 16 principal components (PCs) fitted using the training data. Following the vision-conditioned Diffusion Policy architecture \cite{chi2025diffusion}, we combined the visual features with the tactile PCs, end-effector pose, and gripper width. The policy predicted action chunks containing Cartesian pose and gripper commands, which were executed at $10~\mathrm{Hz}$. We trained each policy for 60 epochs and evaluated three conditions: (i) vision only; (ii) vision with normal-only tactile input (using PCA fitted on the 32 normal channels); and (iii) vision with the full 3-DoF tactile input.

For each of the three conditions, we conducted 25 rollouts at socket locations unseen during training but within the spatial range covered by the training data. A rollout was considered successful if the plug was seated and its indicator light illuminated within $60~\mathrm{s}$. 
% Table~\ref{tab:plug_results} reports the number and rate of successful rollouts.

\begin{table}[b!]
\centering
\caption{Plug-insertion performance at unseen locations.}
\label{tab:plug_results}
\scriptsize
\renewcommand{\arraystretch}{1.05}
\setlength{\tabcolsep}{5pt}
\begin{tabular}{lccc}
\toprule
\textbf{Observation} &
\textbf{Successful trials}  &\textbf{Direct inserts}&\textbf{Successful corrections}\\
\midrule
Vision only             & 12/25 (48\%)&5 &7/20 (35\%)\\
+ normal tactile& 16/25 (64\%)&2 &14/23 (61\%)\\
+ 3-DoF tactile& \textbf{22/25 (88\%)}&4 &\textbf{18/21 (86\%)}\\
\bottomrule
\end{tabular}
\end{table}

% Vision alone succeeded in 12 of 25 trials (48\%). 7 were direct insertions, while only 5 recovered after a missed first attempt. Adding normal tactile feedback raised success to 16 of 25 trials (64\%), and full 3-DoF sensing further increased it to 22 of 25 trials (88\%). These results suggest that shear helped the policy recognize misalignment and perform corrections. 
Success improved monotonically across the three conditions (Table~\ref{tab:plug_results}), while the shear readings helped with correction rather than first-attempt alignment. 

Fig.~\ref{fig:demo}(B.ii) shows a recorded, typical contact sequence during a successful rollout with full 3-DoF tactile sensing, learned solely from the demonstrations. In most cases, a direct insertion failed, leading to a  finding hole step (1). The recorded rotational shear pattern was consistent with the longer earth pin contacting first and causing the plug to experience pivotal force. During this step, the learned policy led to small movements being made, until the earth pin is partially engaged with the socket hole. Then, an alignment step (2) prepared the plug for insertion. At the beginning of this step, the concentrated normal and shear forces on the right finger were consistent with asymmetric contact, providing directional information for adjustment. The shear response decreased during the final descent (3) as the aligned plug entered the socket. Once fully seated, the shear field opposed the insertion direction, signaling the policy to stop (4). The tactile response returned toward baseline after release (5). These signals captured contact transitions that were occluded from the
cameras and provided state information absent from the normal-only condition. Neither VecTile showed visible damage throughout these contact-rich experiments, despite a material cost below 4~USD per sensing surface.

\section{Conclusion}
We introduced VecTile, a flexible $4\times4$ tactile array that recovers three-axis force at each taxel using passive silicone nubs to convert shear into a measurable pressure difference across four piezoresistive quadrants. The design requires no specialized fabrication equipment beyond a 3D printer and a hobby-grade vacuum chamber, and costs under 35~USD per unit (under 15~USD with a cheaper microcontroller). Finite element analysis showed that normal and shear loading remain mechanically separable and that nub height primarily controls shear gain; hardware experiments confirmed this, with the 3.0~mm design reaching 0.39~N normal and 0.16~N shear RMSE over 1--5~N normal and up to 1.5~N shear. A decoder fitted in a flat configuration transferred to $R=85$~mm convex and concave mounts with an average RMSE increase of 0.06~N, and a handful of minutes manual calibration against an F/T reference proved sufficient for practical use. Two demonstrations showed the value of the shear channel: a shear-triggered reflex stopped a scalpel before puncturing an inflated balloon in 20 of 20 trials, where normal force alone did not, and adding 3-DoF tactile input to a Diffusion Policy raised plug-insertion success from 48\% (vision only) and 64\% (normal force only) to 88\%.

Several limitations remain. The resistive layer loses sensitivity above roughly 5~N per taxel, bounding the force range, and we did not yet characterize hysteresis, drift, or long-term durability. Future work will therefore examine whether other readily available elastomers or alternative nub geometries can extend this range and sharpen the shear response, as well as quantify time-dependent effects systematically. Furthermore, the general transducer mechanism and fabrication procedure allow for easy form factor extension. We have already started designing a VecTile version that can be easily cut into a desired footprint via scissors. Finally, we plan to benchmark VecTile against commercial tactile sensors and to test whether its readings can be mapped into sensor-agnostic tactile representations such as those proposed in \cite{hou2025unitac,hou2026unitac}. 

\section*{Acknowledgment}
% We thank [redacted] and [redacted] of [redacted] for loaning us the ATI Mini40 F/T sensor. We also thank [redacted] for advice on the downstream task diffusion policy.
We thank Etienne Burdet and Yihan Liu for lending us the ATI Mini40 F/T sensor. We also thank Qiyang Yan and Xinyun Chi for discussions on the diffusion policy.

\bibliographystyle{IEEEtran}
\bibliography{references}

\end{document}